%% file: main.tex
\documentclass[runningheads]{llncs}
\usepackage{graphicx}
\usepackage{subfiles}
\usepackage{cite}
\usepackage{graphicx}
\usepackage{cancel}
\usepackage{amsmath,amssymb}
\usepackage{booktabs}
\usepackage[hyphens]{url}
\usepackage[misc]{ifsym}
\usepackage[ruled,vlined]{algorithm2e}
\SetKw{KwEstep}{E-step}
\SetKw{KwSstep}{S-step}
\SetKw{KwMstep}{M-step}
\SetAlCapNameFnt{\scriptsize}
\SetAlCapFnt{\scriptsize}
\usepackage{floatrow}
\newfloatcommand{capbtabbox}{table}[][\FBwidth]

\usepackage{blindtext}

\usepackage{makecell}

\begin{document}

\title{Deep Adaptive Multi-Intention Inverse Reinforcement Learning}

\titlerunning{Deep Adaptive Multi-Intention IRL}

\author{Ariyan Bighashdel\textsuperscript{(\Letter)} \and 
Panagiotis Meletis \and
Pavol Jancura \and
Gijs Dubbelman}

\authorrunning{A. Bighashdel et al.}

\institute{Eindhoven University of Technology, 5612 AZ Eindhoven, The Netherlands \\
\email{$\{$a.bighashdel, p.c.meletis, p.jancura, g.dubbelman$\}$@tue.nl}}

\toctitle{Deep Adaptive Multi-Intention Inverse Reinforcement Learning}
\tocauthor{Ariyan~Bighashdel, Panagiotis~Meletis, Pavol~Jancura, \mbox{and Gijs~Dubbelman}}

\maketitle              % typeset the header of the contribution

% \tableofcontents
%%%%%%%%%%%%%%%%%%%%%%%%%%%%%%%%%%%%%%%%%%%%%%%%%%%%%%%%%%%%%%%%%%%%%%%%%%%%%%%%
\begin{abstract}
\subfile{Sections/Abstract}
\end{abstract}

%%%%%%%%%%%%%%%%%%%%%%%%%%%%%%%%%%%%%%%%%%%%%%%%%%%%%%%%%%%%%%%%%%%%%%%%%%%%%%%%
\section{Introduction}
\label{sec:Introduction}
\subfile{Sections/Introduction}

%%%%%%%%%%%%%%%%%%%%%%%%%%%%%%%%%%%%%%%%%%%%%%%%%%%%%%%%%%%%%%%%%%%%%%%%%%%%%%%%
\section{RELATED WORKS}
\label{sec:Related works}
\subfile{Sections/RelatedWorks}

%%%%%%%%%%%%%%%%%%%%%%%%%%%%%%%%%%%%%%%%%%%%%%%%%%%%%%%%%%%%%%%%%%%%%%%%%%%%%%%%
\section{PROBLEM DEFINITION}
\label{sec:Problem}
\subfile{Sections/Problem}

%%%%%%%%%%%%%%%%%%%%%%%%%%%%%%%%%%%%%%%%%%%%%%%%%%%%%%%%%%%%%%%%%%%%%%%%%%%%%%%%
\section{APPROACH}
\label{sec:Approach}
\subfile{Sections/Approach}

%%%%%%%%%%%%%%%%%%%%%%%%%%%%%%%%%%%%%%%%%%%%%%%%%%%%%%%%%%%%%%%%%%%%%%%%%%%%%%%%
\section{EXPERIMENTAL RESULTS}
\label{sec:Experimental Results}
\subfile{Sections/ExperimentalResults}

%%%%%%%%%%%%%%%%%%%%%%%%%%%%%%%%%%%%%%%%%%%%%%%%%%%%%%%%%%%%%%%%%%%%%%%%%%%%%%%%
\section{CONCLUSIONS}
\label{sec:Conclusions}
\subfile{Sections/Conclusions}

\section*{Acknowledgments}
This research has received funding from ECSEL JU in collaboration with the European Union’s 2020 Framework Programme and National Authorities, under grant agreement no. 783190.

%
%
%


% \bibliography{references}
\clearpage

\section*{Appendix A}
\subfile{Sections/AppendixA}
\section*{Appendix B}
\subfile{Sections/AppendixB}
\section*{Appendix C}
\subfile{Sections/AppendixC}

\end{document}

%% file: Sections/Abstract.tex
This paper presents a deep Inverse Reinforcement Learning (IRL) framework that can learn an \textit{a priori} unknown number of nonlinear reward functions from unlabeled experts' demonstrations. For this purpose, we employ the tools from Dirichlet processes and propose an adaptive approach to simultaneously account for both complex and unknown number of reward functions. Using the conditional maximum entropy principle, we model the experts' multi-intention behaviors as a mixture of latent intention distributions and derive two algorithms to estimate the parameters of the deep reward network along with the number of experts' intentions from unlabeled demonstrations. The proposed algorithms are evaluated on three benchmarks, two of which have been specifically extended in this study for multi-intention IRL, and compared with well-known baselines. We demonstrate through several experiments the advantages of our algorithms over the existing approaches and the benefits of online inferring, rather than fixing beforehand, the number of expert's intentions.


%% file: Sections/Introduction.tex
The task of learning from demonstrations (LfD) lies in the heart of many artificial intelligence applications \cite{10.1007/978-3-319-46227-1_33,10.1007/978-3-030-67667-4_32}. By observing the expert's behavior, an agent learns a mapping between world states and actions. This so-called \textit{policy} enables the agent to select and perform an action, given the current world state. Despite the fact that this policy can be directly learned from expert's behaviors, inferring the \textit{reward function} underlying the policy is generally considered the most succinct, robust, and transferable methodology for the LfD task \cite{abbeel2007application}. Inferring the reward function, which is the objective of Inverse Reinforcement Learning (IRL), is often very challenging in real-world scenarios. The demonstrations come from multiple experts who can have different intentions, and their behaviors are consequently not well modeled with a single reward function. Therefore, in this study, we research and extend the concept of \textit{mixture of conditional maximum entropy models} and propose a deep IRL framework to infer an \textit{a priori} unknown number of reward functions from experts' demonstrations without intention labels.\par

Standard IRL can be described as the problem of extracting a reward function, which is consistent with the observed behaviors \cite{russell1998learning}. Obtaining the exact reward function is an ill-posed problem, since many different reward functions can explain the same observed behaviors \cite{ng2000algorithms,ziebart2008maximum}. Ziebart et al. \cite{ziebart2008maximum} tackled this ambiguity by employing the principle of maximum entropy \cite{jaynes1957information}. 
The principle states that the probability distribution, which best represents the current state of knowledge, is the one with the largest entropy \cite{jaynes1957information}. Therefore, Ziebart et al. \cite{ziebart2008maximum} chose the distribution with maximal information entropy to model the experts' behaviors. The maximum entropy IRL has been widely employed in various applications \cite{8463196,8793649}. However, this method suffers from a strong assumption that the experts have one single intention in all demonstrations. In this study, we explore the principle of the mixture of maximum entropy models \cite{pavlov2003mixtures} that inherits the advantages of maximum entropy principle, while at the same time is capable of modeling multi-intention behaviors. \par

In many real-world applications, the demonstrations are often collected from multiple experts whose intentions are potentially different from each other \cite{choi2012nonparametric,almingol2013learning,almingol2015learning,babes2011apprenticeship}. This leads to multiple reward functions, which is in direct contradiction with the single reward assumption in traditional IRL. To address this problem, Babes et al. \cite{babes2011apprenticeship} proposed a clustering-IRL scheme where the class of each demonstration is jointly learned via the respective reward function. Despite the recovery of multiple reward functions, the number of clusters in this method is assumed to be known \textit{a priori}. To overcome this assumption, Choi et al. \cite{choi2012nonparametric} presented a non-parametric Bayesian approach using the Dirichlet Process Mixture (DPM) to infer an unknown number of reward functions from unlabeled demonstrations. However, the proposed method is formulated based on the assumption that the reward functions are formed by a linear combination of a set of world state features. In our work, we do not make this assumption on linearity and model the reward functions using deep neural networks. \par

DPM is a stochastic process in the Bayesian non-parametric framework that deals with mixture models with a countably infinite number of mixture components \cite{neal2000markov}. In general, full Bayesian inference in DPM models is not feasible, and instead, approximate methods like Monte-Carlo Markov chain (MCMC) \cite{li2019tutorial,andrieu2003introduction} and variational inference \cite{blei2004variational} are employed. When deep neural networks are involved in DPM (e.g. deep nonlinear reward functions in IRL), approximates methods may not be able to scale with high dimensional parameter spaces. MCMC sampling methods are shown to be slow in convergence \cite{papamarkou2019challenges,blei2004variational} and variational inference algorithms suffer from restrictions in the distribution family of the observable data, as well as various truncation assumptions for the variational distribution to yield a finite dimensional representation \cite{DBLP:conf/iclr/NalisnickS17,echraibi2020variational}. These limitations apparently make approximate Bayesian inference methods inapplicable for DPM models with deep neural networks. Apart from that, the algorithms for maximum likelihood estimations like standard EM are no longer tractable when dealing with DPM models. The main reason is that the number of mixture components exponentially grows with non-zero probabilities, and after some iterations, the Expectation-step would be no longer available in a closed-form. However, inspired by two variants of EM algorithms that cope with infeasible Expectation-step \cite{wei1990monte,Celeux1985TheSA}, we propose two solutions in which the Expectation-step is either estimated numerically with sampling (based on Monte Carlo EM \cite{wei1990monte}) or computed analytically and then replaced with a sample from it (based on stochastic EM \cite{Celeux1985TheSA}).\par

This study's main contribution is to develop an IRL framework where one can benefit from the strength of 1) maximum entropy principle, 2) deep nonlinear reward functions, and 3) account for an unknown number of experts' intentions. To the best of our knowledge, we are the first to present an approach that can combine all these three capabilities. \par
In our proposed framework, the experts' behavioral distribution is modeled as a mixture of conditional maximum entropy models. The reward functions are parameterized as a deep reward network, consisting of two parts: 1) a base reward model, and 2) an adaptively growing set of intention-specific reward models. The base reward model takes as input the state features and outputs a set of reward features shared in all intention-specific reward models. The intention-specific reward models take the reward features and output the rewards for the respective expert's intention. A novel adaptive approach, based on the concept of the Chinese Restaurant Process (CRP), is proposed to infer the number of experts' intentions from unlabeled demonstrations. To train the framework, we propose and compare two novel EM algorithms. One is based on stochastic EM and the other on Monte Carlo EM. In Section \ref{sec:Problem}, this problem of multi-intention IRL is defined, following our two novel EM algorithms in Section \ref{sec:Approach}. The results are evaluated on three available simulated benchmarks, two of which are extended in this paper for multi-intention IRL, and compared with two baselines \cite{babes2011apprenticeship,choi2012nonparametric}. These experimental results are reported in Section \ref{sec:Experimental Results} and Section \ref{sec:Conclusions} is devoted to conclusions. The source code to reproduce the experiments is publicly available\footnote{\url{https://github.com/tue-mps/damiirl}}.

%% file: Sections/RelatedWorks.tex
In the past decades, a number of studies have addressed the problem of multi-intention IRL. A comparison of various methods for multi-intention IRL, together with our approach, is depicted in Table \ref{tab:table1}. \par
In an early work, Dimitrakakis and Rothkopf \cite{dimitrakakis2011bayesian} formulated the problem of learning from unlabeled demonstrations as a multi-task learning problem. By generalizing the Bayesian IRL approach of Ramachandran and Amir \cite{ramachandran2007bayesian}, they assumed that each observed trajectory is responsible for one specific reward function, all of which shares a common prior. The same approach has also been employed by Noothigattu et al. \cite{noothigattu2020inverse}, who assumed that each expert's reward function is a random permutation of one sharing reward function.
Babes et al. \cite{babes2011apprenticeship} took a different approach and addressed the problem as a clustering task with IRL. They proposed an EM approach that clusters the observed trajectories by inferring the rewards function for each cluster. Using maximum likelihood, they estimated the reward parameters for each cluster. \par
The main limitation in EM clustering approach is that the number of clusters has to be specified as an input parameter \cite{babes2011apprenticeship,nguyen2015inverse}. To overcome this assumption, Choi and Kim \cite{choi2012nonparametric} employed a non-parametric Bayesian approach via the DPM model. Using MCMC sampler, they were able to infer an unknown number of reward functions, which are linear combinations of state features. Other authors have also employed the same methodology in the literature \cite{rajasekaran2017inverse,michini2012bayesian,almingol2013learning}. \par
All above methods are developed on the basis of model-based reinforcement learning (RL), in which the model of the environment is assumed to be known. In the past few years, a couple of approximate, model-free methods have been developed for IRL with multiple reward functions \cite{hausman2017multi,li2017infogail,hsiao2019learning,lin2018acgail}. Such methods aimed to solve large-scale problems by approximating the Bellman optimality equation with model-free RL. 

In this study, we constrain ourselves to model-based RL and propose a multi-intention IRL approach to infer an unknown number of experts' intentions and corresponding nonlinear reward functions from unlabeled demonstrations. 
\begin{table}
	\centering
	\caption{Comparison of proposed models for multi-intention IRL.}
	\resizebox{\textwidth}{!}{%
	\begin{tabular}{l|cc|ccc}
		\toprule
		\multicolumn{1}{c}{}   &
		\multicolumn{2}{c}{\bf{Type}}  &
		\multicolumn{3}{c}{\bf{Features}}   \\
		\cmidrule(r){2-3}
		\cmidrule(r){4-6}
		\bf{Models}     & \makecell{Model\\based}    & \makecell{Model\\free}  & \makecell{Unlabeled\\demonstrations} & \makecell{Unknown \#\\intentions} & \makecell{Non-linear\\reward fun.}\\
		\midrule
		Dimitrakakis and Rothkopf \cite{dimitrakakis2011bayesian}& \checkmark &   & \checkmark      &     & \\
		Babes et al. \cite{babes2011apprenticeship}& \checkmark &       & \checkmark  &   & \\
		Nguyen et al. \cite{nguyen2015inverse}& \checkmark &   & \checkmark  &   &  \\
		 Choi and Kim \cite{choi2012nonparametric} & \checkmark &   & \checkmark      & \checkmark  & \\
		Rajasekaran et al. \cite{rajasekaran2017inverse}& \checkmark &       & \checkmark  & \checkmark  & \\
		Li et al. \cite{li2017infogail} &      & \checkmark  & \checkmark  &  & \checkmark\\
		Hausman et al. \cite{hausman2017multi} &     & \checkmark   & \checkmark  &   & \checkmark\\
		Lin and Zhang \cite{lin2018acgail} &     &  \checkmark  &   &   & \checkmark\\
		Hsiao et al. \cite{hsiao2019learning} &      & \checkmark  & \checkmark  &  & \checkmark\\
		\bottomrule
		\bf{Ours} &   \checkmark   &   & \checkmark  & \checkmark  & \checkmark\\
		\bottomrule
        
	\end{tabular}
	}
	\label{tab:table1}
\end{table}
\vspace{-12pt}

%% file: Sections/Problem.tex
In this section, the problem of multi-intention IRL is defined. To facilitate the flow, we first formalize the multi-intention RL problem. For both problems, we follow the conventional modelling of the environment as a Markov Decision Process (MDP). A finite state MDP in a multi-intention RL problem is a tuple $(S,A,T,\gamma,b_0,$ $R_1,R_2,...,R_K)$ where $S$ is the state space , $A$ is the action space, $T:S\times A \times S \rightarrow [0,1]$ is the transition probability function, $\gamma \in [0,1)$ is the discount factor, $b_0(s)$ is the probability of staring in state $s$, and $R_k:S   \rightarrow {\mathbb{R}}$ is the $k^{th}$ reward function with $K$ to be the total number of intentions. A policy is a mapping function $\pi_k:S \rightarrow A \;\;\forall k \in \{1,2,...,K\}$. The value of policy $\pi_k$ with respect to the $k^{th}$ reward function is the expected discounted reward for following the policy and is defined as $V_{R_k}^{\pi} = {\mathbb{E}} [\sum_t \gamma^t R_k(s_t)|b_0] $. The optimal policy ($\pi_k^*$) for the $k^{th}$ reward function  is the policy that maximizes the value function for all states and satisfies the respective Bellman optimality equation \cite{sutton2018reinforcement}.\par
In multi-intention IRL, the context of this study, a finite-state MDP{\textbackslash}R is a tuple $(S,A,T,\gamma,b_0,$ $\pmb{\tau}^1,\pmb{\tau}^2,...,\pmb{\tau}^M)$ where $\pmb{\tau}^m$ is the $m^{th}$ demonstration and $M$ is the total number of demonstrations. In this work, it is assumed that there is a total of $K$ intentions, each of which corresponds to one reward function, so that $\pmb{\tau}^m$ with length $T_\tau$ is generated from the optimal policy ($\pi_k^*$) of the $k^{th}$ reward function. It is further assumed that the demonstrations are without intention labels, i.e. they are \textit{unlabeled}. Therefore, the goal is to infer the number of intentions $K$ and the respective reward function of each intention.
In the next section, we model the experts' behaviors as a mixture of conditional maximum entropy models, parameterize the reward functions via deep neural networks, and propose a novel approach to infer an unknown number of experts' intentions from unlabeled demonstrations. 

%% file: Sections/Approach.tex
In the proposed framework for multi-intention IRL, the experts' behavioral distribution is modeled as a mixture of conditional maximum entropy models. The Mixture of conditional maximum entropy models is a generalization of standard maximum entropy formulation for cases where the data distributions arise from a mixture of simpler underlying latent distributions \cite{pavlov2003mixtures}. According to this principal, a mixture of conditional maximum entropy models is a promising candidate to justify the multi-intention behaviors of the experts. The experts' behaviors with the $k^{th}$ intention is defined via a conditional maximum entropy distribution:
\begin{equation}
	\begin{split}
		p(\pmb{\tau}|\eta_k=1, \Psi) = exp(R_k(\pmb{\tau},\Psi_k))/Z_k,
	\end{split}
\end{equation}
\begin{figure}
	\centering
	\includegraphics[width=0.47\columnwidth, keepaspectratio]{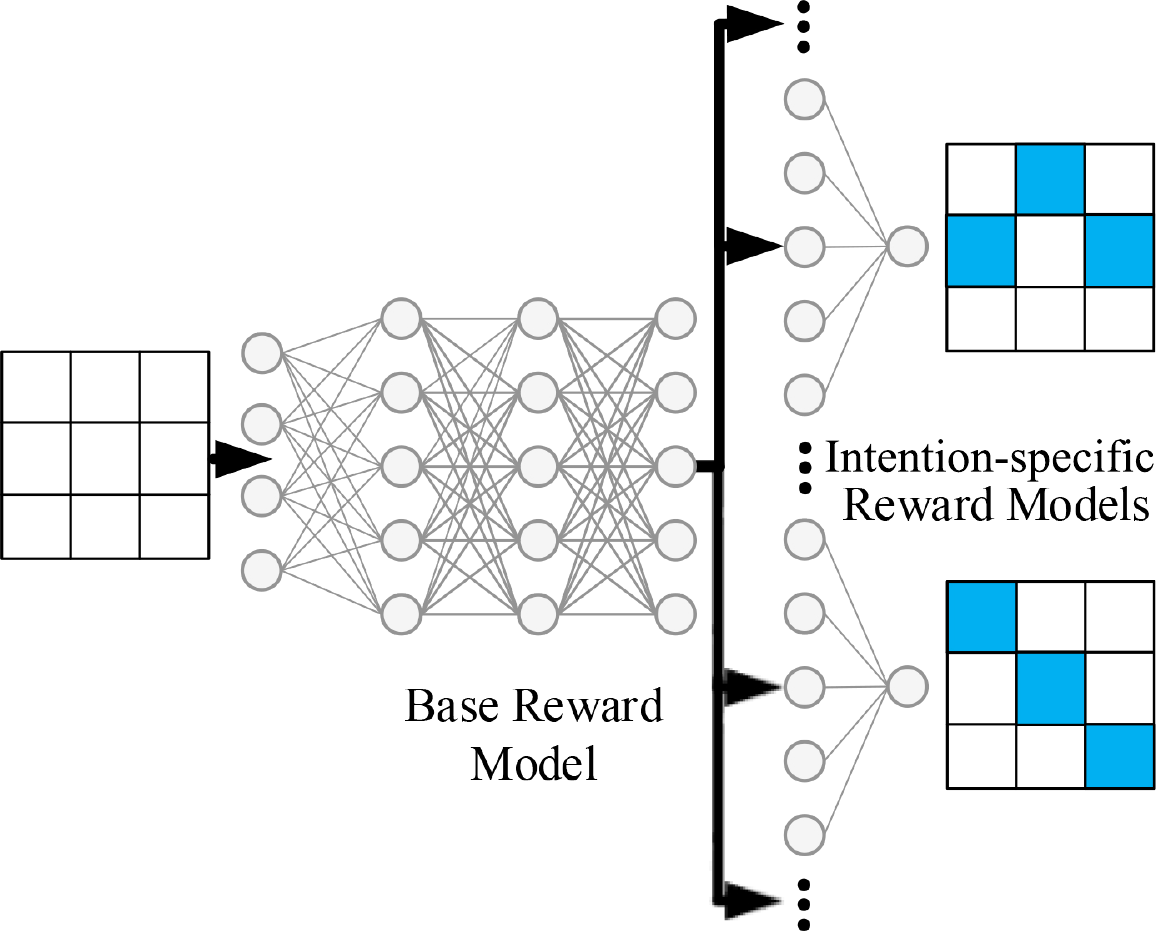}
	\caption{Schematics of deep reward network.}
	\label{fig:exp6}
\end{figure}
where $\pmb{\eta}=\{\eta_1,\eta_2,...,\eta_K |\forall \eta_k \in \{0,1\}, \sum_{k=1}^K \eta_k = 1\}$ is the latent intention vector, $R_k(\pmb{\tau},\Psi_k) = \sum_{s \in \pmb{\tau}}R_k(s,\Psi_k)$ is the reward of the trajectory with respect to the $k^{th}$ reward function with $R_k(s,\Psi_k)$ as the state reward value, and $Z_k$ is the $k^{th}$ partition function. \par
We define the $k^{th}$ reward function as: $R_k(s,\Psi_k) = R_{{\Psi}_k}(\pmb{f}_s)$, where $R_{{\Psi}_k}$ is a deep neural network with finite set of parameters $\Psi_k = \{\Theta_0,\Theta_k\}$ which consists of a base reward model $R_{{\Theta_0}}$ and an intention-specific reward model $R_{{\Theta}_k}$ (See Fig. \ref{fig:exp6}). The base reward model with finite set of parameters $\Theta_0$ takes the state feature vector $\pmb{f}_s$ and outputs the state reward feature vector $\pmb{r}_s$: $\pmb{r}_s = R_{{\Theta_0}}(\pmb{f}_s)$. The state reward feature vector $\pmb{r}_s$ that is produced by the base reward model is input to all intention-specific reward models. The $k^{th}$ intention-specific reward model with finite set of parameters $\Theta_k$, takes the state reward feature vector $\pmb{r}_s$ and outputs the state reward value: $R_k(s, \Psi_k) = R_{\Theta_k}(\pmb{r}_s)$. Therefore the total set of reward parameters is $\Psi=\{\Theta_0,\Theta_1,...,\Theta_K \}$. The reward of the trajectory $\pmb{\tau}$ with respect to the $k^{th}$ reward function can be further obtained as: $R_{k}(\pmb{\tau},\Psi_k) = \pmb{\mu}(\pmb{\tau}) ^\intercal \pmb{R}_{{\Psi}_k}(\pmb{\tau})$,
where $\pmb{\mu}(\pmb{\tau})$ is the expected State Visitation Frequency (SVF) vector for trajectory $\pmb{\tau}$ and $\pmb{R}_{{\Psi}_k}(\pmb{\tau})=\{R_{{\Psi}_k}(\pmb{f}_s)|\forall s \in S \}$ is the vector of reward values of all states with respect to the $k^th$ reward function.\par
In order to infer the number of intentions $K$, we propose an adaptive approach in which the number of intentions adaptively changes whenever a trajectory is visited/re-visited. For this purpose, at each iteration we first assume to have $M-1$ demonstrated trajectories $\{\pmb{\tau}^1,\pmb{\tau}^2,...,\pmb{\tau}^{m-1},$ $\pmb{\tau}^{m+1},...,\pmb{\tau}^M\}$ that are already assigned to $K$ intentions with known latent intention vectors $\pmb{H}^{-m} = \{\pmb{\eta}^1,\pmb{\eta}^2,...,\pmb{\eta}^{m-1},\pmb{\eta}^{m+1},...,\pmb{\eta}^{M} \}$. Then, we visit/re-visit a demonstrated trajectory $\pmb{\tau}^{m}$ and the task is to obtain the latent intention vector $\pmb{\eta}^{m}$, which can be assigned to a new intention $K+1$, and update the reward parameters $\Psi$. As emphasized before, our work aims to develop a method in which K, the number of intentions, is a priori unknown and can, in theory, be arbitrarily large. Now we define the predictive distribution for the trajectory $\pmb{\tau}^{m}$ as a mixture of conditional maximum entropy models:
\begin{equation}
\begin{split}
    p(\pmb{\tau}^m|\pmb{H}^{-m},\Psi) =  \sum_{k=1}^{K+1} p(\pmb{\tau}^m|\eta^m_k=1,\Psi)p(\eta^m_k=1|\pmb{H}^{-m})
\end{split}
\end{equation}
where $p(\eta^m_k=1|\pmb{H}^{-m})$ is the prior intention assignment for trajectory $\pmb{\tau}^m$, given all other latent intention vectors. In the case of $K$ intentions, we define a multinomial prior distribution over all latent intention vectors $\pmb{H}=\{\pmb{H}^{-m},\pmb{\eta}^m\}$:
\begin{equation}
\begin{split}
    p(\pmb{H}|\pmb{\phi}) &= \prod_{k=1}^K \phi_k ^ {M_k}
\end{split}
\end{equation}
where $M_k$ is the number of trajectories with intention $k$ and $\pmb{{\phi}}$ is the vector of mixing coefficients $\pmb{\phi}=\{\phi_1, \phi_2,...\phi_K\}$ with Dirichlet prior  distribution $p(\pmb{\phi}) = \text{Dir}(\alpha/K)$,
where $\alpha$ is the concentration parameter. As  $K\rightarrow\infty$ the main problematic parameters are the mixing coefficients. Marginalizing out the mixing coefficients and separating the latent intention vector for $m^{th}$ trajectory yield (see Appendix A for full derivation):
\begin{equation}
	\label{eq:priorCRP}
	\begin{split}
		& p(\eta^m_k=1|\pmb{H}^{-m}) = \frac{M_{k}^{-m}}{M-1+\alpha}\\
		& p(\eta_{K+1}^{m}=1|\pmb{H}^{-m}) = \frac{\alpha}{M-1+\alpha}
	\end{split}
\end{equation}
where $M_{k}^{-m}$ is the number of trajectories assigned to intention $k$ excluding the $m^{th}$ trajectory, $p(\eta_k^{m}=1|\pmb{H}^{-m})$ is the prior probability of assigning the new trajectory $m$ to intention $k \in \{1,2,...,K\}$, and $p(\eta_{K+1}^{m}=1|\pmb{H}^{-m})$ is the prior probability of assigning the new trajectory $m$ to intention $K+1$. Equation (\ref{eq:priorCRP}) is known as the CRP representation for DPM \cite{neal2000markov}. Considering the exchangeability property \cite{gershman2012tutorial}, the following optimization problem is defined:
\begin{equation}
\begin{split}
    \mathop{\max}_{\Psi}L^m(\Psi) = &\log \sum_{k=1}^{K+1} p(\pmb{\tau}^m|\eta^m_k=1,\Psi)p(\eta^m_k=1|\pmb{H}^{-m})\,\,\,\,\,\,\,\,\,\,\forall m \in \{1,2,...,M\}
\end{split}
\end{equation}
The parameters $\Psi$ can be estimated via Expectation Maximization (EM) \cite{bishop2006pattern}. Differentiating $L^m(\Psi)$ with respect to $\psi \in \Psi$ yields the following E-step and M-step (see Appendix B for full derivation):
\subsubsection{E-step}
Evaluation of the posterior distribution over the latent intention vector $\forall k \in \{1,2,...,K\}$:
\begin{equation}
\label{E-step1}
\begin{split}
    \gamma^m_k & =\frac{M_k^{-m}\prod_{t=0}^{T_\tau-1} \pi_{k}(a_t|s_t)}{\alpha \prod_{t=0}^{T_\tau-1} \pi_{K+1}(a_t|s_t) + \sum_{\hat{k}=1}^{K} M_k^{-m}\prod_{t=0}^{T_\tau-1} \pi_{\hat{k}}(a_t|s_t)}
\end{split}
\end{equation}
and for $k = K+1$:
\begin{equation}
\label{E-step2}
\begin{split}
    \gamma^m_k & =\frac{\alpha\prod_{t=0}^{T_\tau-1} \pi_{k}(a_t|s_t)}{\alpha \prod_{t=0}^{T_\tau-1} \pi_{K+1}(a_t|s_t) + \sum_{\hat{k}=1}^{K} M_k^{-m}\prod_{t=0}^{T_\tau-1} \pi_{\hat{k}}(a_t|s_t)}
\end{split}
\end{equation}
where we have defined $\gamma^m_k = p(\eta^m_k=1|\pmb{\tau}^m,\pmb{H}^{-m},\Psi)$.
\subsubsection{M-step}
update of the parameter value $\psi \in \Psi$ with gradient of:
\begin{equation}
\label{M-step}
\begin{split}
    \nabla_{\psi}L(\Psi) = \sum_{k=1}^{K+1}\gamma^m_k(\pmb{\mu}(\pmb{\tau}^m) -{{\mathbb{E}}}_{p(\pmb{\tau}|\eta_k=1,\Psi)}[\pmb{\mu}(\pmb{\tau})]) ^\intercal \frac{d{{\pmb{R}}_{\Psi_k}(\pmb{\tau})}}{d\psi}
\end{split}
\end{equation}
where ${{\mathbb{E}}}_{p(\pmb{\tau}|\eta_k=1,\Psi)}[\pmb{\mu}(\pmb{\tau})]$ is the expected SVF vector under the parameterized reward function $R_{\Psi_k}$ \cite{ziebart2008maximum}.\par
When $K$ approaches infinity, the EM algorithm is no longer tractable since the number of mixture components exponentially grows with non-zero probabilities. As a result, after some iterations, the E-step would be no longer available in a closed-form. We propose two solutions for estimation of the reward parameters which are inspired by stochastic and Monte Carlo EM algorithms. Both proposed solutions are deeply evaluated and compared with in Section \ref{sec:Experimental Results}.

\begin{algorithm}
	\caption{Adaptive multi-intention IRL based on stochastic EM}
	\label{alg:SEM}
	\SetAlgoLined
	Initialize $K$, $\Theta_0, \Theta_1,\Theta_2,...,\Theta_K$, $M_1,M_2,...,M_K$ \;
	\While{$iteration<MaxIter$}{
		Solve for $\pi_{1},\pi_{2},...,\pi_{K}$\;
		\For{$m=1$ \KwTo $M$}{
			Initialize $\Theta_{K+1}$ and solve for $\pi_{{K+1}}$\;
			\KwEstep{Obtain $\gamma^m_k$ $\forall k \in \{1,2,...,K,K+1\}$}\;
			\KwSstep{Sample $\eta^m_k \sim \gamma^m_k$}  \;
			\If{$\eta^m_{K+1}=1$}{
				$K=K+1$\;
			}
			Remove $K_{u}$ unoccupied intentions: \
			$K=K-K_u$\;
			Update $M_1,M_2,...,M_K$\;
		
    		\KwMstep{Update $\psi \in \{\Theta_0, \Theta_1,\Theta_2,...,\Theta_K\}$ by (\ref{M-step})}\;
    	}
	}
\end{algorithm}
\vspace{-12pt}

\subsection{First solution with stochastic expectation maximization}
Stochastic EM, introduces a stochastic step (S-step) after the E-step that represents the full expectation with a single sample \cite{Celeux1985TheSA}. Alg. \ref{alg:SEM} presents the summary of the first solution to multi-intention IRL via stochastic EM algorithm when the number of intentions is no longer known. \par 
Given (\ref{E-step1}) and (\ref{E-step2}), first the posterior distribution over the latent intention vector $\pmb{\eta}^m$ for trajectory $\pmb{\tau}^m \in \{\pmb{\tau}^1,\pmb{\tau}^2,...,\pmb{\tau}^M \}$ is obtained. Then, the full expectation is estimated with a sample $\pmb{\eta}^m$ from the posterior distribution. Finally, the reward parameters are updated via (\ref{M-step}).

\subsection{Second solution with Monte Carlo expectation maximization}
The Monte Carlo EM algorithm is a modification of the EM algorithm where the expectation in the E-step is computed numerically via Monte Carlo simulations \cite{wei1990monte}. As indicated, Alg. \ref{alg:SEM} relies on the full posterior distribution which can be time-consuming. Therefore, another solution for multi-intention IRL is presented in which the E-step is performed through Metropolis-Hastings sampler (see Alg. \ref{alg:MCEM} for the summary). \par
First, a new intention assignment for $m^{th}$ trajectory, $\pmb{\eta}^{*m}$, is sampled from the prior distribution of (\ref{eq:priorCRP}), then $\pmb{\eta}^{m}=\pmb{\eta}^{*m}$ is set with the acceptance probability of $min\{1, \frac{p(\pmb{\tau}^m|\pmb{\eta}^{*m},\Psi)}{p(\pmb{\tau}^m|\pmb{\eta}^{m},\Psi)} \}$
where (see Appendix C for full derivation):
\begin{equation}
	\begin{split}
		\frac{p(\pmb{\tau}^m|\eta^{*m}_{k^*}=1,\Psi)}{p(\pmb{\tau}^m|\eta^{m}_k=1,\Psi)} = \frac{\prod_{t=1}^{T_\tau}\pi_{{k^*}}(a_t^m|s_t^m)}{\prod_{t=1}^{T_\tau}\pi_{{k}}(a_t^m|s_t^m)}
	\end{split}
\end{equation}
with $k \in \{1,2,...,K\}$ and $k^* \in \{1,2,...,K,K+1\}$.
\begin{algorithm}
	\caption{Adaptive multi-intention IRL based on Monte Carlo EM}
	\label{alg:MCEM}
	\SetAlgoLined
	Initialize $K$, $\Theta_0, \Theta_1,\Theta_2,...,\Theta_K$, $M_1,M_2,...,M_K$ \;
	\While{$iteration<MaxIter$}{
		Solve for $\pi_{1},\pi_{2},...,\pi_{K}$\;
		\For{$m=1$ \KwTo $M$}{
			Obtain $p(\pmb{\eta}^m|\pmb{H}^{-m},\alpha)$\;
			Sample $\pmb{\eta}^{*m} \sim p(\pmb{\eta}^m|\pmb{H}^{-m},\alpha)$\;
			\If{$\eta^{*m}_{K+1}=1$}{
				Initialize $\Theta_{K+1}$ and solve for $\pi_{{K+1}}$\;
			}
			\KwEstep{Assign $\pmb{\eta}^{*m} \rightarrow \pmb{\eta}^{m}$ by probability of $min\{1, \frac{p(\pmb{\tau}^m|\pmb{\eta}^{*m},\Psi)}{p(\pmb{\tau}^m|\pmb{\eta}^{m},\Psi)} \}$}\;
			\If{$\eta^m_{K+1}=1$}{
				$K=K+1$\;
			}
			Remove $K_{u}$ unoccupied intentions: \
			$K=K-K_u$\;
			Update $M_1,M_2,...,M_K$\;
		
    		\KwMstep{Update $\psi \in \{\Theta_0, \Theta_1,\Theta_2,...,\Theta_K\}$ by (\ref{M-step})}\;
    	}
	}
\end{algorithm}

%% file: Sections/ExperimentalResults.tex
In this section, we evaluate the performance of our proposed methods through several experiments with three goals: 1) to show the advantages of our methods in comparison with the baselines in environments with both linear and non-linear rewards, 2) to demonstrate the advantages of adaptively inferring the number of intentions rather than predefining a fixed number, and 3) to depict the strengths and weaknesses of our proposed algorithms with respect to each other.

\subsection{Benchmarks}
In order to deeply compare the performances of various models, the experiments are conducted on three different environments: GridWorld, Multi-intention ObjectWorld, and Multi-intention BinaryWorld. Variants of all three environments have been widely employed in IRL literature \cite{levine2011nonlinear,wulfmeier2015maximum}.  \par
GridWorld \cite{choi2012nonparametric} is a $8 \times 8$ environment with 64 states and four actions per state with 20\% probability of moving randomly. The grids are partitioned into non-overlapping regions of size $2 \times 2$, and the feature function is defined by a binary indicator function for each region. Three reward functions are generated with linear combinations of state features and reward weights which are sampled to have a non-zero value with the probability of 0.2. The main idea behind using this environment is to compare all the models in aspects other than their capability of handling linear/non-linear reward functions.\par
Multi-intention ObjectWorld (M-ObjectWorld) is our extension of ObjectWorld \cite{levine2011nonlinear} for multi-intention IRL. ObjectWorld is a $32\times 32$ grid of states with five actions per state with a 30\% chance of moving in a different random direction. The objects with two different inner and outer colors are randomly placed, and the binary state features are obtained based on the Euclidean distance to the nearest object with a specific inner or outer color. Unlike ObjectWorld, M-ObjectWorld has six different reward functions, each of which corresponds to one intention. The intentions are defined for each cell based on three rules: 1) within 3 cells of outer color one and within 2 cells of outer color two, 2) Just within 3 cells of outer color one, and 3) everywhere else (see Table 2). Due to the large number of irrelevant features and the nonlinearity of the reward rules, the environment is challenging for methods that learn linear reward functions. Fig. 2 (top three) shows a $8 \times 8$ zoom-in of M-ObjectWorld with three reward functions and respective optimal policies.\par
Multi-intention BinaryWorld (M-BinaryWorld) is our extension of BinaryWorld \cite{wulfmeier2015maximum} for multi-intention IRL. Similarly, BinaryWorld has $32\times 32$ states, five actions per state with a 30\% chance of moving in a different random direction. But every state is randomly occupied with one of the two-color objects. The feature vector for each state consequently consists of a binary vector, encoding the color of each object in $3 \times 3$ neighborhood. Similar to M-ObjectWorld, six different intentions can be defined for each cell of M-BinaryWorld based on three rules: 1) four neighboring cells have color one, 2) five neighboring cells have color one, and 3) everything else (see Table 2). Since in M-BinaryWorld the reward depends on a higher representation for the basic features, the environment is arguably more challenging than the previous ones. Therefore, most of the experiments are carried in this environment. Fig. 2 (bottom three) shows a $8 \times 8$ zoom-in of M-BinaryWorld with three different reward functions and policies.\par
In order to assess the generalizability of the models, the experiments are also conducted on \textit{transferred} environments. In transferred environments, the learned reward functions are re-evaluated on new randomized environments.

\begin{figure}
\begin{floatrow}
\ffigbox[][][]{\caption{$8 \times 8$ zoom-ins of M-ObjectWorld (top three) and M-BinaryWorld (bottom three) with three reward functions.}}
        {\includegraphics[width=1\columnwidth, keepaspectratio]{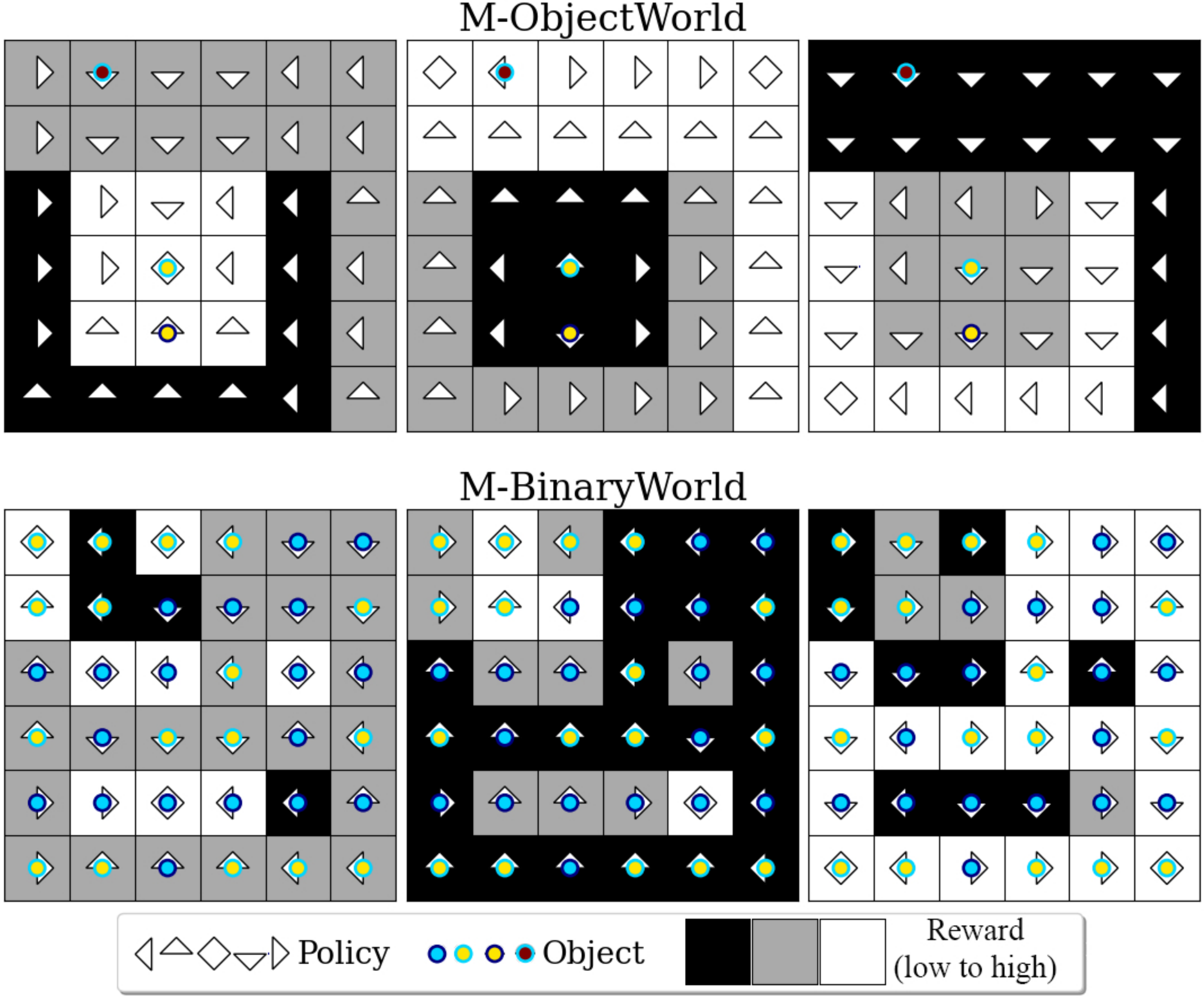}}
\capbtabbox{%
	\begin{tabular}{c|ccc}
		\toprule
		\multicolumn{1}{c}{}   &
		\multicolumn{3}{c}{\bf{Reward rule}}  \\
		\cmidrule(r){2-4}
		\bf{Intention}     & \bf{1}    & \bf{2}   & \bf{3} \\
		\midrule
		A& +5 & -10  & 0      \\
		B& -10 & 0      & +5  \\
		C& 0     & +5  & -10  \\
		D& -10 & +5  & 0      \\
		E& +5 & 0      & -10  \\
		F& 0     & -10  & +5  \\
		\bottomrule
	\end{tabular}
}{%
  \caption{Reward values in M-ObjectWorld and M-BinaryWorld}%
}
\end{floatrow}
\end{figure}

\subsection{Models}
In this study, we compare our methods with existing approaches that can handle IRL with multiple intentions and constrain the experiments to model-based methods. The following models are evaluated on the benchmarks: 
\begin{itemize}
  \item EM-MLIRL($K$), proposed by Babes et al. \cite{babes2011apprenticeship}. This method requires the number of experts' intentions $K$ to be known. To research the influence on setting $K$ for this method, we use $K \in \{2,3,4\}$.
  \item DPM-BIRL, a non-parametric multi-intention IRL method proposed by Choi and Kim \cite{choi2012nonparametric}.
  \item SEM-MIIRL, our proposed solution based on stochastic EM.
  \item MCEM-MIIRL, our proposed solution based on Monte Carlo EM.
  \item $K$EM-MIIRL, a simplified variant of our approach where the concentration parameter is zero and the number of intentions are fixed to $K \in \{2,5\}$.
\end{itemize}

\subsection{Metric}
Following the same convention used in \cite{choi2012nonparametric}, the imitation performance is evaluated by the average of expected value difference (EVD). The EVD measures the performance difference between the expert’s optimal policy and the optimal policy induced by the learned reward function. For $m \in \{1,2,...,M\}$, $\text{EVD} = |V^{\Tilde{\pi}^m}_{\Tilde{R}^m}-V^{\pi^m}_{\Tilde{R}^m}|$, where $\Tilde{\pi}^m$ and $\Tilde{R}^m$ are the true policy and reward function for $m^{th}$ demonstration, respectively, and $\pi^m$ is the predicted policy under the predicted reward function demonstration. In all experiments, a lower average-EVD corresponds to better imitation performance.

\subsection{Implementations details}
In our experiments, we employed a fully connected neural network with five hidden layers of dimension 256 and a rectified linear unit for the base reward model, and a set of linear functions represents the intention-specific reward models. The reward network is trained for 200 epochs using Adam \cite{kingma2014adam} with a fixed learning rate of 0.001. For easing the reproducibility of our work, the source code is shared with the community at https://github.com/tue-mps/damiirl.
\begin{figure}
	\centering
	\includegraphics[width=1\textwidth, keepaspectratio]{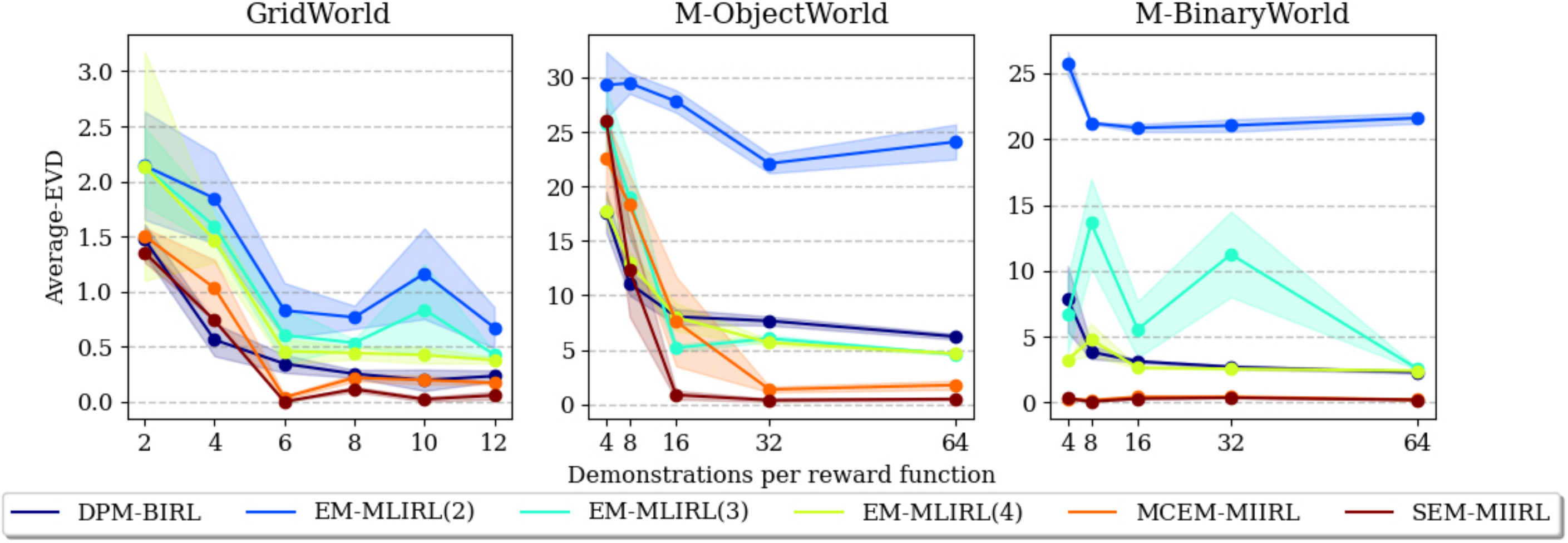}
	\caption{Imitation performance in comparison with the baselines. Lower average-EVD is better.}
	\label{fig:exp1}
\end{figure}
\begin{figure}
	\centering
	\includegraphics[width=0.76\columnwidth, keepaspectratio]{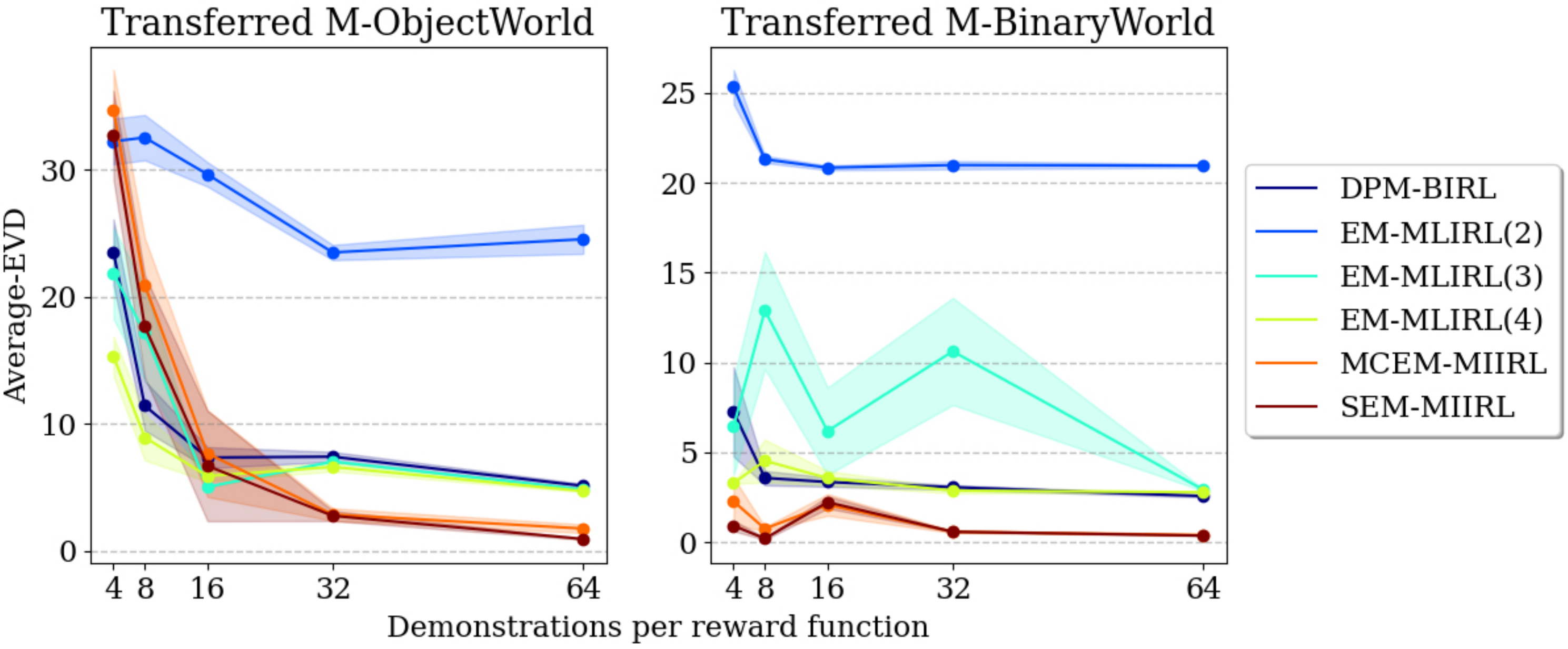}
	\caption{Imitation performance in comparison with the baselines in transferred environments. Lower average-EVD is better.}
	\label{fig:exp1_tf}
\end{figure}
\begin{figure}
	\centering
	\includegraphics[width=0.76\columnwidth, keepaspectratio]{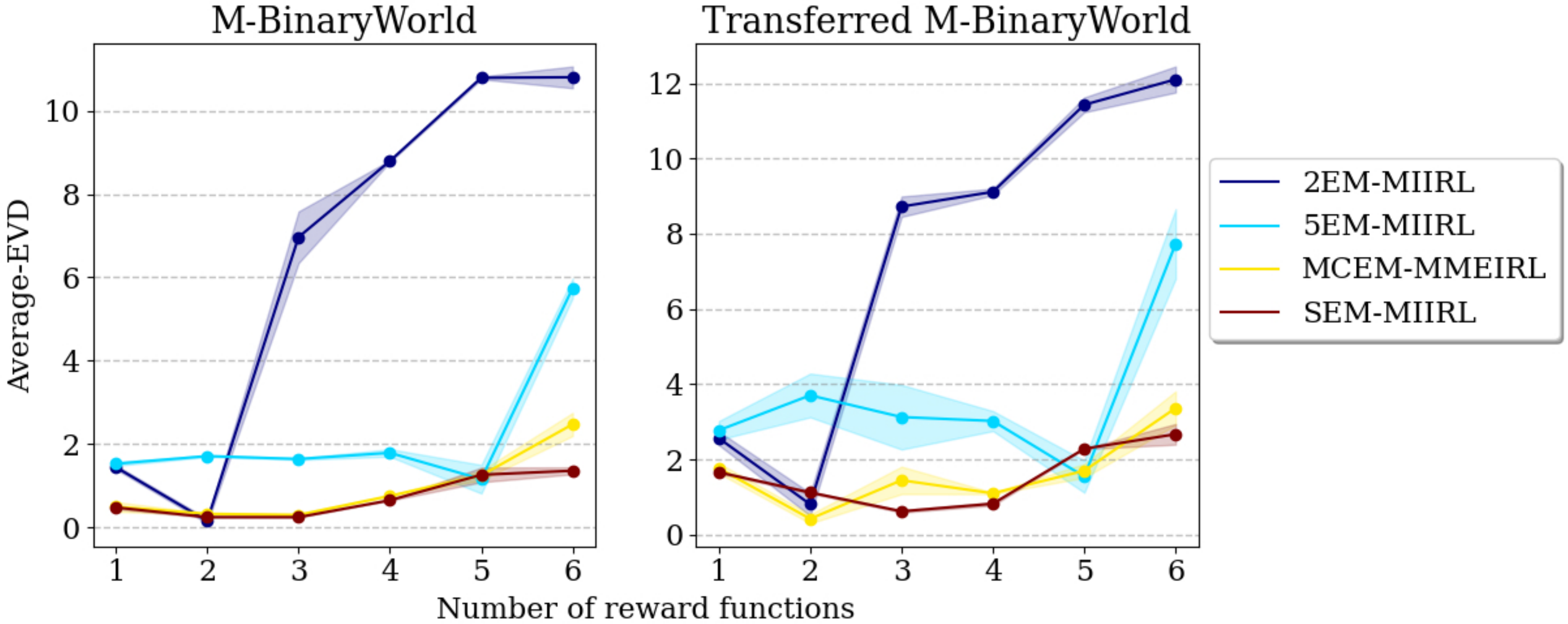}
	\caption{Effects of overestimating/underestimating vs inferring the number of reward functions in original (left) and transferred (right) M-BinaryWorlds. Lower average-EVD is better}
	\label{fig:exp2}
\end{figure}
\subsection{Results}
Each experiment is repeated for 6 times with different random environments, and the results are shown in the form of means (lines) and standard errors (shadings). The demonstration length for GridWorld is fixed to 40 time-steps and for both M-ObjectWorld and M-BinaryWorld is 8 time-steps. \par
\begin{figure}
	\centering
	\includegraphics[width=0.76\columnwidth, keepaspectratio]{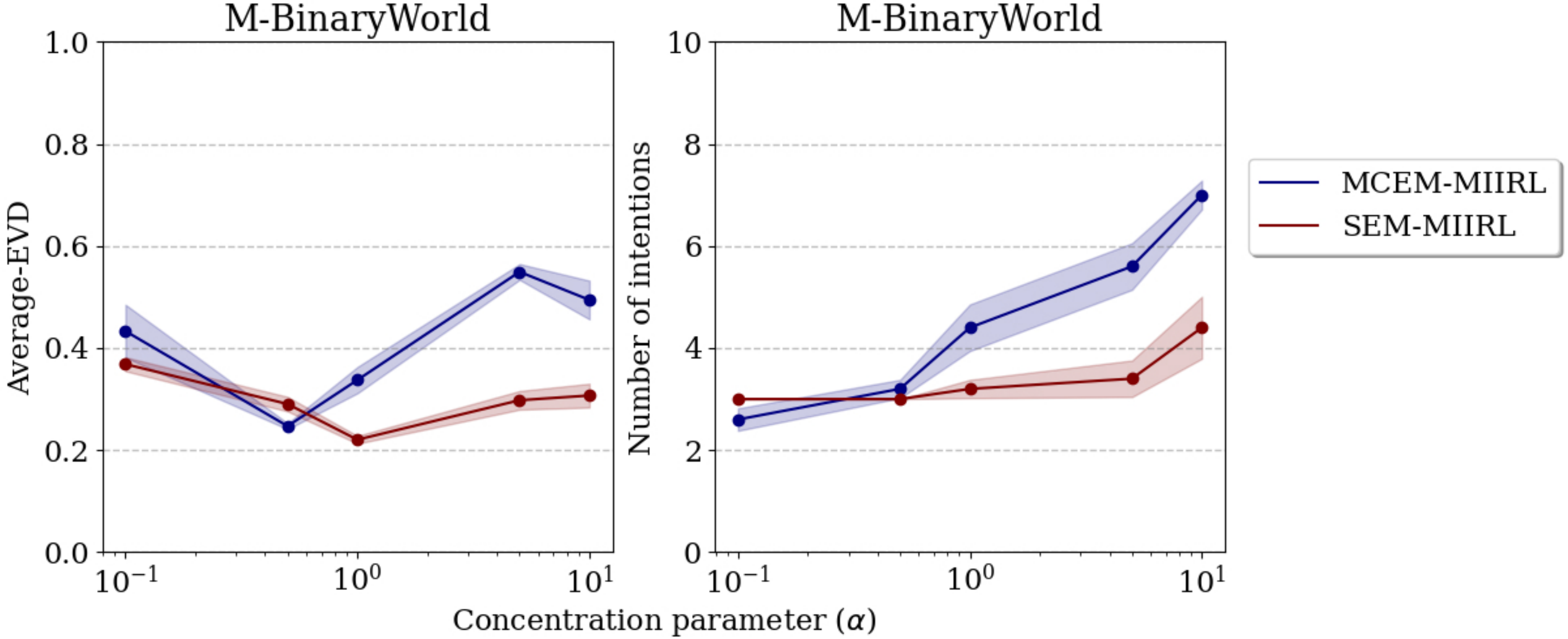}
	\caption{Effects of $\alpha$ on Average-EVD (left) and number of predicted intentions (right). Lower average-EVD is better}
	\label{fig:exp3}
\end{figure}
Fig. \ref{fig:exp1} and Fig. \ref{fig:exp1_tf} show the imitation performances of our SEM-MIIRL and MCEM-MIIRL in comparison with two baselines, EM-MLIRL($K$) and  DPM-BIRL, for varying number of demonstrations per reward function in original and transferred environments, respectively. Each expert is assigned to one out of three reward functions (intentions A, B, and C in M-ObjectWorld and M-BinaryWorld) and the concentration parameter is set to one. The results show clearly that our methods achieve significant lower average-EVD errors when compared to existing methods, especially in nonlinear environments of M-ObjectWorld and M-BinaryWorld, with SEM-MIIRL slightly outperforming MCEM-MIIRL \par
\begin{figure}
	\centering
	\includegraphics[width=0.76\columnwidth, keepaspectratio]{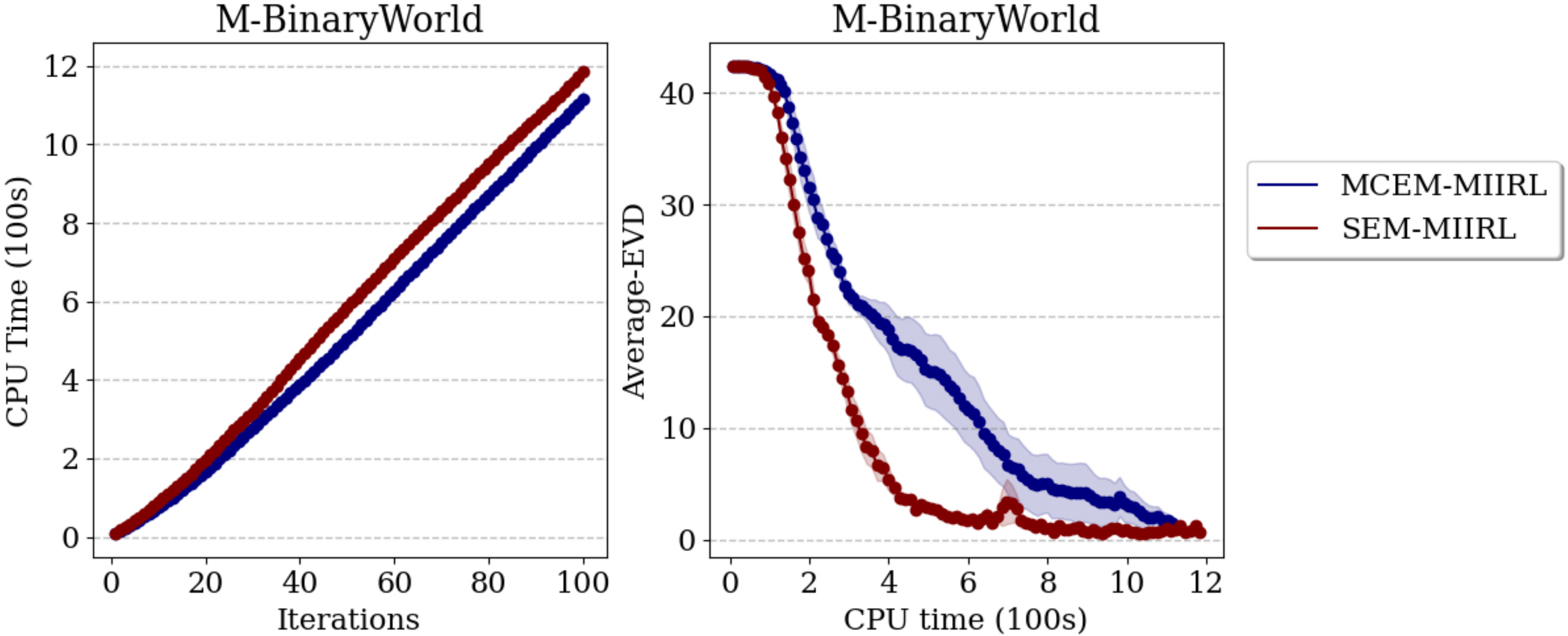}
	\caption{Execution time (right) and Convergence (left). Lower average-EVD is better.}
	\label{fig:exp4}
\end{figure}
To address the importance of inferring the number of intentions, we have compared the performances of our SEM-MIIRL and MCEM-MIIRL with two simplified variants, 2EM-MIIRL and 5EM-MIIRL, where the concentration parameter is set to zero and the number of intentions is fixed and equal to 2 and 5, respectively. Fig. \ref{fig:exp2} shows the results of these comparisons for a varying number of true reward functions from one to six (from intention: \{A\} to \{A, B, C, D, E, F\}) in both original and transferred M-BinaryWorld. The number of demonstrations is fixed to $16$ per reward function and $\alpha=1$ for both SEM-MIIRL and MCEM-MIIRL. As depicted, overestimation and underestimation of the number of reward functions, as happens frequently in both 2EM-MIIRL and 5EM-MIIRL, deteriorate the imitation performance. This while the adaptability in 
SEM-MIIRL and MCEM-MIIRL yields to less sensitivity with changes in the number of true reward functions.\par
Further experiments are conducted to deeply assess and compare MCEM-MIIRL and SEM-MIIRL. Fig. \ref{fig:exp3} depicts the effects of the concentration parameter on both Average-EVD and number of predicted intentions. The number of demonstrations is fixed to $16$ per reward function and intentions are \{A, B, C\}. As shown, the best value for the concentration parameter is between 0.5 to 1, with lower values leading to higher Average-EVD and lower number of predicted intentions, while higher values result in higher Average-EVD and higher number of predicted intentions for both MCEM-MIIRL and SEM-MIIRL. The final experiment is devoted to the convergence behavior of MCEM-MIIRL and SEM-MIIRL. The number of demonstrations is again fixed to $16$ per reward function, intentions are \{A, B, C\} and the concentration parameter is set to $1$. As shown in Fig. \ref{fig:exp4} (left image), the per-iteration execution time of MCEM-MIIRL is lower than SEM-MIIRL. The main reason is that SEM-MIIRL evaluates the posterior distribution over all latent intentions. However, this extra operation guarantees faster converges of SEM-MIIRL, making it overall the more efficient than MCEM-MIIRL as can be seen in Fig.  \ref{fig:exp4} (right image).

%% file: Sections/Conclusions.tex
We proposed an inverse reinforcement learning framework to recover complex reward functions by observing experts whose behaviors originate from an unknown number of intentions. We presented two algorithms that are able to consistently recover multiple, highly nonlinear reward functions and whose benefits were pointed out through a set of experiments. For this, we extended two complex benchmarks for multi-intention IRL in which our algorithms distinctly outperformed the baselines. We also demonstrated the importance of inferring rather than underestimating or overestimating the number of experts' intentions\par
Having shown the benefits of our approach in inferring the unknown number of experts' intention from a collection of demonstrations via model-based RL, we aim to extend the same approach in model-free environments by employing approximate RL methods.

%% file: Sections/AppendixA.tex
We assume that we have $M-1$ demonstrated trajectories with a set of known latent intention vectors  $\pmb{H}^{-m} = \{\pmb{\eta}^1,\pmb{\eta}^2,...,\pmb{\eta}^{m-1},\pmb{\eta}^{m+1},...,\pmb{\eta}^{M} \}$ with $K$ intentions. Then, we have a new  demonstrated trajectory $\pmb{\tau}^{m}$ and the task is to obtain the latent intention vector $\pmb{\eta}^{m}$, which can be a new intention $K+1$, and update the reward parameters $\Psi$. We are willing to consider growing/infinite number of intentions. \par
In the case of $K$ intentions, we define a Categorical prior distribution over $\pmb{H}=\{\pmb{H}^{-m},\pmb{\eta}^m\}$:

\begin{equation}
\begin{split}
    p(\pmb{H}|\pmb{\phi}) &= \prod_{m=1}^M \text{Cat}(\pmb{{\phi}})\\
    &= \prod_{k=1}^K \phi_k ^ {M_k}
\end{split}
\end{equation}
where $M_k$ is the number of trajectories with intention $k$ and $\pmb{{\phi}}$ is the vector of mixing coefficients $\pmb{\phi}=\{\phi_1, \phi_2,...\phi_K\}$ with prior distribution of:
\begin{equation}
\begin{split}
    p(\pmb{\phi}) &= \text{Dir}(\alpha/K)\\
    &= \frac{\Gamma(\alpha)}{\Gamma(\alpha/K)^K}\prod_{k=1}^K\pi_k^{\alpha/K-1}
\end{split}
\end{equation}
where  $\alpha$ is the concentration parameter. The main problematic variable as  $K\rightarrow\infty$ are the mixing coefficients. We marginalize out $\pmb{\phi}$:
\begin{equation}
\begin{split}
    p(\pmb{H})&=\int p(\pmb{H}|\pmb{\phi})p(\pmb{\phi})\\
    &=\frac{\Gamma(\alpha)}{\Gamma(M+\alpha)}\prod_{k=1}^K\frac{\Gamma(M_k+\alpha/K)}{\Gamma(\alpha/K)}
\end{split}
\end{equation}
Given that:
\begin{equation}
\begin{split}
    p(\pmb{H}) = p(\pmb{\eta}^m|\pmb{H}^{-m})p(\pmb{H}^{-m})
\end{split}
\end{equation}
we can define the conditional prior over $\pmb{\eta}^m=\{\eta^m_1,\eta^m_2,...,\eta^m_K\}$ as:
\begin{equation}
\begin{split}
    p(\eta^m_k=1|\pmb{H}^{-m}) = \frac{M_k^{-m}+\alpha/K}{M-1+\alpha}
\end{split}
\end{equation}
where $M_k^{-m}$ is the number of trajectories with intention $k$ excluding $\pmb{\tau}^m$. By letting $K \rightarrow \infty$, we reach:
\begin{equation}
    \label{eq:crpp1}
	\begin{split}
		p(\eta^m_k=1|\pmb{H}^{-m}) = \frac{M_k^{-m}}{M-1+\alpha}
	\end{split}
\end{equation}
where $	p(\eta^m_k=1|\pmb{H}^{-m})$ is the prior probability of assigning the trajectory $\pmb{\tau}^{m}$ to intention $k \in \{1,2,...,K\}$. Since:
\begin{equation}
	\begin{split}
		\sum_{k=1}^K p(\eta^m_k=1|\pmb{H}^{-m}) = \frac{M-1}{M-1+\alpha} \neq 1 
	\end{split}
\end{equation}
we define $	p(\eta^m_{K+1}=1|\pmb{H}^{-m})$ as the prior probability of assigning the trajectory $\pmb{\tau}^{m}$ to intention $k+1$:
\begin{equation}
    \label{eq:crpp2}
	\begin{split}
	p(\eta^m_{K+1}=1|\pmb{H}^{-m}) &= 1 - \frac{M-1}{M-1+\alpha}\\
	& = \frac{\alpha}{M-1+\alpha}
	\end{split}
\end{equation}
Equations (\ref{eq:crpp1}) and (\ref{eq:crpp2}) are known as Chinese Restaurant Process \cite{li2019tutorial}. 

%% file: Sections/AppendixB.tex
Given the predictive distribution for $m^{th}$ trajectory:
\begin{equation}
\begin{split}
    p(\pmb{\tau}^m|\pmb{H}^{-m},\Psi) =  \sum_{k=1}^{K+1} p(\pmb{\tau}^m|\eta^m_k=1,\Psi)p(\eta^m_k=1|\pmb{H}^{-m})
\end{split}
\end{equation}
the following optimization problem can be defined $\forall m \in \{1,2,...,M\}$ by employing the exchangeability property \cite{gershman2012tutorial}:
\begin{equation}
\begin{split}
    \mathop{\max}_{\Psi}L^m(\Psi) = \log \sum_{k=1}^{K+1} p(\pmb{\tau}^m|\eta^m_k=1,\Psi)p(\eta^m_k=1|\pmb{H}^{-m})
\end{split}
\end{equation}
The parameters $\Psi$ can be estimated via Expectation Maximization (EM) \cite{bishop2006pattern}. Differentiating the log-likelihood function $L(\Psi)$ with respect to $\psi \in \Psi$ yields:
\begin{equation}
\label{eq2}
\begin{split}
    \nabla_{\psi}L^m & =  \frac{\nabla_{\psi}\sum_{k=1}^{K+1} p(\pmb{\tau}^m|\eta^m_k=1,\Psi)p(\eta^m_k=1|\pmb{H}^{-m})}{\sum_{\hat{k}} p(\pmb{\tau}^m|\eta^m_k=1,\Psi)p(\eta^m_k=1|\pmb{H}^{-m})}\\
    &= \sum_{k=1}^{K+1}\frac{\nabla_{\psi} p(\pmb{\tau}^m|\eta^m_k=1,\Psi)p(\eta^m_k=1|\pmb{H}^{-m})}{\sum_{\hat{k}} p(\pmb{\tau}^m|\eta^m_k=1,\Psi)p(\eta^m_k=1|\pmb{H}^{-m})}
\end{split}
\end{equation}
A standard trick in setting up the EM procedure is to introduce the posterior distribution over the latent intention vector $\pmb{\eta}^m$ \cite{bishop2006pattern}:
\begin{equation}
\label{eq3}
\begin{split}
    \gamma^m_k =  p(\eta^m_k=1|\pmb{\tau}^m,\pmb{H}^{-m},\Psi) & = \frac{p(\pmb{\tau}^m,\eta^m_k=1|\pmb{H}^{-m},\Psi)}{\sum_{\hat{k}=1}^{K+1} p(\pmb{\tau}^m,\eta^m_{\hat{k}}=1|\pmb{H}^{-m},\Psi)} \\
    &= \frac{p(\pmb{\tau}^m|\eta^m_k=1,\Psi)p(\eta^m_k=1|\pmb{H}^{-m})}{\sum_{\hat{k}=1}^{K+1} p(\pmb{\tau}^m|\eta^m_{\hat{k}}=1,\Psi)p(\eta^m_{\hat{k}}=1|\pmb{H}^{-m})}
\end{split}
\end{equation}
Now the term under summation in (\ref{eq2}) can be written as::
\begin{equation}
\label{eq4}
\begin{split}
    &\frac{\nabla_{\psi} p(\pmb{\tau}^m|\eta^m_k=1,\Psi)p(\eta^m_k=1|\pmb{H}^{-m})}{\sum_{\hat{k}=1}^{K+1} p(\pmb{\tau}^m|\eta^m_k=1,\Psi)p(\eta^m_k=1|\pmb{H}^{-m})}\\
    &=\frac{p(\pmb{\tau}^m|\eta^m_k=1,\Psi)p(\eta^m_k=1|\pmb{H}^{-m})}{\sum_{\hat{k}=1}^{K+1} p(\pmb{\tau}^m|\eta^m_k=1,\Psi)p(\eta^m_k=1|\pmb{H}^{-m})}\frac{\nabla_{\psi} p(\pmb{\tau}^m|\eta^m_k=1,\Psi)p(\eta^m_k=1|\pmb{H}^{-m})}{p(\pmb{\tau}^m|\eta^m_k=1,\Psi)p(\eta^m_k=1|\pmb{H}^{-m})}\\
     &=\gamma^m_k \frac{\nabla_{\psi} p(\pmb{\tau}^m|\eta^m_k=1,\Psi)p(\eta^m_k=1|\pmb{H}^{-m})}{p(\pmb{\tau}^m|\eta^m_k=1,\Psi)p(\eta^m_k=1|\pmb{H}^{-m})}\\
     &=\gamma^m_k\nabla_{\psi}\log p(\pmb{\tau}^m|\eta^m_k=1,\Psi)p(\eta^m_k=1|\pmb{H}^{-m})
\end{split}
\end{equation}
Performing the differentiation of the second term in (\ref{eq4}) yields:
\begin{equation}
\begin{split}
     \nabla_{\psi}&\log p(\pmb{\tau}^m|\eta^m_k=1,\Psi)p(\eta^m_k=1|\pmb{H}^{-m}) \\ 
     &=\nabla_{\psi}\log p(\pmb{\tau}^m|\eta^m_k=1,\Psi)+ \cancelto{0}{\nabla_{\psi}\log p(\eta^m_k=1|\pmb{H}^{-m})}\\
     & =\nabla_{\psi}\log(\frac{\exp(R_{k}(\pmb{\tau}^m,\psi_k))}{Z(k)})\\
     & =\nabla_{\psi}({R_{k}(\pmb{\tau}^m,\psi_k)} - \log Z(k)) \\
     & =\nabla_{\psi}({R_{k}(\pmb{\tau}^m,\psi_k)} - \log \sum_\tau \exp({R_{k}(\pmb{\tau},\psi_k)}))\\
     & = \frac{d{R_{k}(\pmb{\tau}^m,\psi_k)}}{d\psi} - \frac{\sum_\tau \exp({R_{k}(\pmb{\tau},\psi_k)})){\frac{d{R_{k}(\pmb{\tau},\psi_k)}}{d\psi}}}{\sum_\tau \exp({R_{k}(\pmb{\tau},\psi_k)}))} \\
    & = \frac{d{R_{k}(\pmb{\tau}^m,\psi_k)}}{d\psi}  - \sum_{\tau}{p(\pmb{\tau}|\eta_k=1,\Psi)}{\frac{d{R_{k}(\pmb{\tau},\psi_k)}}{d\psi}}\\
    & = (\pmb{\mu}(\pmb{\tau}^m) -{{\mathbb{E}}}_{p(\pmb{\tau}|\eta_k=1,\Psi)}[\pmb{\mu}(\pmb{\tau})]) ^\intercal \frac{d{{\pmb{R}}_{\Psi_k}(\pmb{\tau})}}{d\psi}
\end{split}
\end{equation}
Therefore (\ref{eq2}) results in: 
\begin{equation}
\begin{split}
    \nabla_{\psi}L = \sum_{k=1}^{K+1}\gamma^m_k(\pmb{\mu}(\pmb{\tau}^m) -{{\mathbb{E}}}_{p(\pmb{\tau}|\eta_k=1,\Psi)}[\pmb{\mu}(\pmb{\tau})]) ^\intercal \frac{d{{\pmb{R}}_{\Psi_k}(\pmb{\tau})}}{d\psi}
\end{split}
\end{equation}
which is known as the M-step. The posterior distribution over the latent intention vector $\pmb{\eta}^m$ can be obtained as:
\begin{equation}
\begin{split}
    \gamma^m_k & =\frac{p(\pmb{\tau}^m|\eta^m_k=1,\Psi)p(\eta^m_k=1|\pmb{H}^{-m})}{\sum_{\hat{k}=1}^{K+1} p(\pmb{\tau}^m|\eta^m_{\hat{k}}=1,\Psi)p(\eta^m_{\hat{k}}=1|\pmb{H}^{-m})}\\
    & = \frac{b_0(s_0)\prod_{t=0}^{T-1} T(s_{t+1}|s_t,a_t)\pi_{k}(a_t|s_t)p(\eta^m_k=1|\pmb{H}^{-m})}{\sum_{\hat{k}=1}^{K+1}b_0(s_0)\prod_{t=0}^{T-1} T(s_{t+1}|s_t,a_t)\pi_{\hat{k}}(a_t|s_t)p(\eta^m_{\hat{k}}=1|\pmb{H}^{-m})}\\
    & = \frac{\prod_{t=0}^{T-1} \pi_{k}(a_t|s_t)p(\eta^m_k=1|\pmb{H}^{-m})}{\sum_{\hat{k}=1}^{K+1} \prod_{t=0}^{T-1} \pi_{\hat{k}}(a_t|s_t)p(\eta^m_{\hat{k}}=1|\pmb{H}^{-m})}
\end{split}
\end{equation}
Using (\ref{eq:crpp1}) and (\ref{eq:crpp2}) yields $\forall k \in \{1,2,...,K\}$:
\begin{equation}
\begin{split}
    \gamma^m_k & =\frac{M_k^{-m}\prod_{t=0}^{T-1} \pi_{k}(a_t|s_t)}{\alpha \prod_{t=0}^{T-1} \pi_{K+1}(a_t|s_t) + \sum_{\hat{k}=1}^{K} M_k^{-m}\prod_{t=0}^{T-1} \pi_{\hat{k}}(a_t|s_t)}
\end{split}
\end{equation}
and for $K+1$:
\begin{equation}
\begin{split}
    \gamma^m_k & =\frac{\alpha\prod_{t=0}^{T-1} \pi_{k}(a_t|s_t)}{\alpha \prod_{t=0}^{T-1} \pi_{K+1}(a_t|s_t) + \sum_{\hat{k}=1}^{K} M_k^{-m}\prod_{t=0}^{T-1} \pi_{\hat{k}}(a_t|s_t)}
\end{split}
\end{equation}
Which are known as the E-step.

%% file: Sections/AppendixC.tex
The likelihood ratio for the $m^{th}$  trajectory is obtained as:
\begin{equation}
	\begin{split}
		\frac{p(\pmb{\tau}^m|\eta^{*m}_{k^*}=1,\Psi)}{p(\pmb{\tau}^m|\eta^{m}_k=1,\Psi)} &= \frac{b_0(s_0)\prod_{t=1}^{T_\tau}T(s_{t+1}|s_t,a_t)\pi_{{k^*}}(a_t^m|s_t^m)}{b_0(s_0)\prod_{t=1}^{T_\tau}T(s_{t+1}|s_t,a_t)\pi_{{k}}(a_t^m|s_t^m)}\\
		&=  \frac{\prod_{t=1}^{T_\tau}\pi_{{k^*}}(a_t^m|s_t^m)}{\prod_{t=1}^{T_\tau}\pi_{{k}}(a_t^m|s_t^m)}
	\end{split}
\end{equation}
with $k \in \{1,2,...,K\}$ and $k^* \in \{1,2,...,K,K+1\}$.s